\pdfoutput=1
\documentclass[letterpaper]{article} 
\usepackage[preprint]{aaai2027}  
\usepackage[hyphens]{url}  
\usepackage{graphicx} 
\usepackage{natbib}  
\usepackage{caption} 
\usepackage{subcaption}
\usepackage{algorithm}
\usepackage{algorithmic}
\usepackage{array} 
\usepackage[table]{xcolor}
\definecolor{rowblue}{RGB}{220,230,241} 
\usepackage{makecell}
\usepackage{multirow}
\usepackage{pifont}
\usepackage{placeins}
\newcommand{\cmark}{\textcolor{green!60!black}{\ding{51}}}
\newcommand{\xmark}{\textcolor{red!75!black}{\ding{55}}}
\newcommand{\name}{\emph{OmniCAD}}
\newcommand{\tai}{\textcolor{black}}
\usepackage{newfloat}
\usepackage{listings}
\DeclareCaptionStyle{ruled}{labelfont=normalfont,labelsep=colon,strut=off} 
\floatstyle{ruled}
\newfloat{listing}{tb}{lst}{}
\floatname{listing}{Listing}

\usepackage{booktabs}


\title{OmniCAD: A Large-Scale Benchmark for 3D Spatial Reasoning in Robotics Assemblies}

\author{
    Mingjia Wang\textsuperscript{\rm 1}\equalcontrib,
    Taiting Lu\textsuperscript{\rm 2}\equalcontrib,
    Ziwei Dong\textsuperscript{\rm 3},
    Sisong Bei\textsuperscript{\rm 3},
    Jingying Zeng\textsuperscript{\rm 3},
    Runze Liu\textsuperscript{\rm 2},
    \\
    Kaiyuan Lin\textsuperscript{\rm 2},
    Hongxing Pan\textsuperscript{\rm 1},
    Kai Zhang\textsuperscript{\rm 1},
    Yizheng Hou\textsuperscript{\rm 1},
    Yangshoudu Zheng\textsuperscript{\rm 1},
    Chenchen Guo\textsuperscript{\rm 1},
    \\
    Weiyuan Meng\textsuperscript{\rm 1},
    Shubin Lyu\textsuperscript{\rm 1},
    Zhijun Zheng\textsuperscript{\rm 1},
    Dexu Wang\textsuperscript{\rm 1},
    Xinyu Bai\textsuperscript{\rm 1},
    Shurui Qian\textsuperscript{\rm 1},
    \\
    Zhangzixin\textsuperscript{\rm 1},
    Mengyu Pan\textsuperscript{\rm 1},
    Guoliang Shi\textsuperscript{\rm 1},
    Ling Ma\textsuperscript{\rm 1},
    Yifan Yang\textsuperscript{\rm 4},
    Qi He\textsuperscript{\rm 3},
    \\
    Yi-Chao Chen\textsuperscript{\rm 1},
    Yincheng Jin\textsuperscript{\rm 5},
    Sung-Liang Chen\textsuperscript{\rm 1}\corresponding,
    Mahanth Gowda\textsuperscript{\rm 2}\corresponding
}

\affiliations{
    \textsuperscript{\rm 1}Shanghai Jiao Tong University,
    \textsuperscript{\rm 2}Pennsylvania State University,
    \textsuperscript{\rm 3}Independent Researcher,\\
    \textsuperscript{\rm 4}Microsoft Research,
    \textsuperscript{\rm 5}Binghamton University
}

\begin{document}
\raggedbottom
\setlength{\textfloatsep}{6pt plus 1pt minus 1pt}
\setlength{\floatsep}{6pt plus 1pt minus 1pt}
\setlength{\intextsep}{6pt plus 1pt minus 1pt}

\maketitle
\begin{figure*}[t]
    \centering
    \includegraphics[width=\textwidth]
        {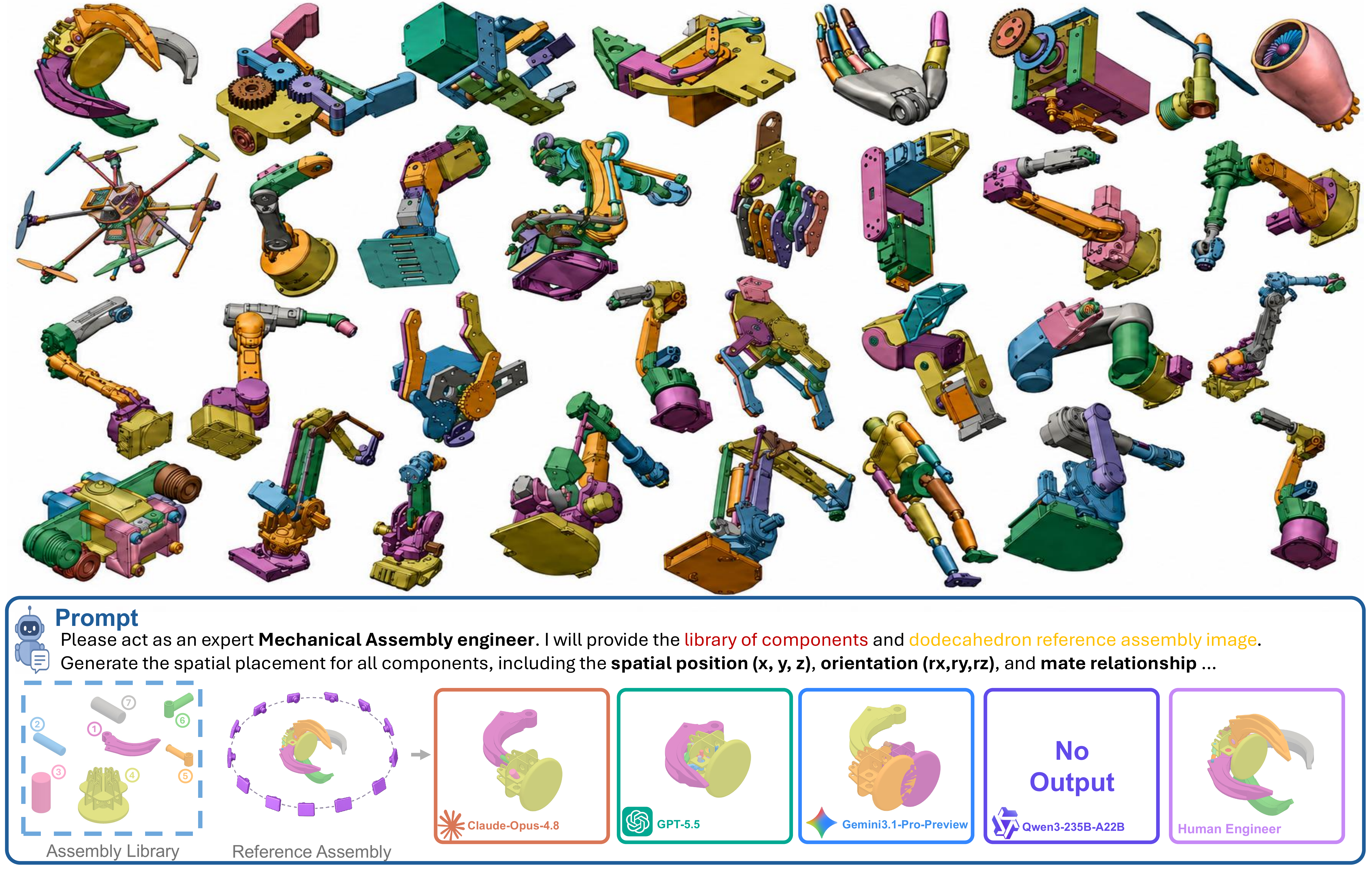}
    \caption{
        VLMs exhibit limited capability in
        assembling 3D mechanical components into physically valid
        configurations. Frequent failures, including incorrect
        component poses, severe part interpenetration, missing or
        duplicated component instances, and inconsistent mating
        relationships, reveal substantial shortcomings in
        assembly-aware spatial and relational reasoning.
    }
    \label{fig:lmm_test}
\end{figure*}

\begin{abstract}
Recent vision-language models (VLMs) have demonstrated strong capabilities in robotic spatial-relation reasoning, object manipulation, environmental perception and task planning.  
However, their ability to reason about complex robotic component assemblies remains largely unexplored, despite being essential for enabling VLMs to move beyond perceiving and manipulating existing environments toward designing, constructing, and reshaping the physical world.
To bridge this gap, we introduce \textbf{\name}, the first large-scale benchmark designed to evaluate VLMs on assembly-aware 3D spatial reasoning across diverse industrial mechanical systems, spanning robotic mechanisms, automotive components, aerospace structures, and agricultural machinery.
\textbf{\name} comprises \tai{25k} industrial mechanical assemblies, each consisting of a corresponding set of component parts, with an average of \tai{12} parts per assembly and spanning \tai{21} distinct types of mate relationships. Each assembly is paired with a ground-truth 3D assembly model constructed and verified by human engineers, together with renderings from 20 distinct viewpoints that provide comprehensive visual coverage of its component geometry, spatial configuration, and mating relationships.
The benchmark comprises three tasks: \textbf{(1) component-level 3D spatial reasoning}, evaluating VLMs on inferring the 3D position and orientation of each component from assembly images; \textbf{(2) part-to-part relational reasoning}, requiring models to identify mating relationships between components and recover the underlying assembly connectivity and constraint structure; and \textbf{(3) tool-augmented agentic assembly reasoning}, evaluating agents that iteratively select informative viewpoints, inspect the resulting visual evidence, and self-correct their predictions of component poses and mating relationships to accomplish tasks (1)–(2).
Our results show that current VLMs struggle with industrial 3D assembly reasoning, often producing incorrect component poses, invalid mating relationships, part interpenetration, and poor scalability to large component libraries. We will open-source the benchmark, evaluation code, and tool interfaces to support research on accurate, physically valid, and scalable 3D assembly.

\end{abstract}

\section{Introduction}\label{intro}

Large vision-language models have substantially advanced the generation, understanding, and automation of digital content, yet their ability to reason about how physical components should be assembled into functional artifacts remains comparatively underexplored. 
This capability is fundamental to real-world engineering, manufacturing, and robotic construction, where system functionality depends not only on individual components but also on their precise spatial arrangement, connectivity, and interactions. 
Well-designed assembly reasoning is therefore a critical step toward extending multimodal intelligence beyond digital content creation to the design and construction of functional physical systems.
However, state-of-the-art approaches \cite{wang2023ikea,tie2025manual2skill,zhang2025manual,li2026assemblybench} to 3D assembly reasoning have primarily focused on household furniture assembly. This focus is largely enabled by the public availability of abundant instruction manuals that provide standardized diagrams and step-by-step assembly procedures. Each assembly in these datasets comprises only a limited number of parts, with clearly illustrated connections and relatively straightforward attachment relationships, compared with practical designs that may involve hundreds or even thousands of components.
Thus, furniture-derived datasets mainly provide a simplified setting for studying assembly reasoning, but do not fully capture the geometric complexity, part ambiguity, and dense mating constraints of industrial mechanical assemblies.

However, real-world industrial assemblies, ranging from robotic systems and automotive mechanisms to aerospace equipment and agricultural machinery, pose substantially greater challenges due to their numerous geometrically complex or visually similar components, dense mating constraints, tight tolerances, and nontrivial insertion and rotational motions.
\tai{
As shown in Table~\ref{tab:assembly_dataset}, existing large-scale
3D assembly datasets with reported statistics contain an average of
approximately 6.7--18.72 component instances per assembly, and most
do not provide explicit mate graphs between components. Moreover,
CAD-centric datasets generally lack standardized multi-view visual
observations, while vision-oriented datasets provide limited
B-Rep geometry and mating-constraint supervision.
}
Meanwhile, vision-language models (VLMs) \cite{liu2023zero,chen2024spatialvlm,niu2026cme,li2026assemblybench} have demonstrated promising capabilities in furniture assembly, 3D spatial reasoning, and fine-grained geometric understanding, suggesting their potential to infer component identities, 6-DoF poses, and part-to-part mating relationships for complex mechanical assemblies from visual observations.
Yet it remains unclear whether these capabilities are sufficient for substantially more demanding industrial assemblies, where models must identify the correct component instances, infer their precise 6-DoF poses and orientations, resolve occlusions and geometric ambiguities, and recover the underlying part-to-part mating relationships that determine a physically valid assembly. 
This raises a critical open question: \textbf{\textit{Given visual references of a target assembly and a library of candidate 3D parts, can current VLMs select the required component instances, including repeated parts, and accurately infer their 6-DoF poses, orientations, and part-to-part mating relationships to recover a physically valid assembly?}}

To investigate this question, we conducted a series of preliminary experiments evaluating the mechanical drawing-to-3D reconstruction capabilities of state-of-the-art general-purpose VLMs including Claude Opus 4.8 \cite{anthropic2026claudeopus48}, GPT-5.5 \cite{openai_gpt55}, Gemini 3.1 Pro Preview \cite{google2026gemini31propreview}, and Qwen3.7-Plus \cite{qwen3_7_plus}.
Their performance was compared with that of reference 3D assembly by experienced mechanical engineers.
As illustrated in Figure \ref{fig:lmm_test}, general-purpose VLMs struggle to accurately identify and select components in complex 3D assemblies, often producing geometrically invalid results with intersecting parts. Their limited input context windows also make it difficult to process assemblies containing large numbers of components and extensive geometric information.

\begin{figure*}[t]
    \centering
    \includegraphics[width=\textwidth,trim=20 10 20 20,clip]
        {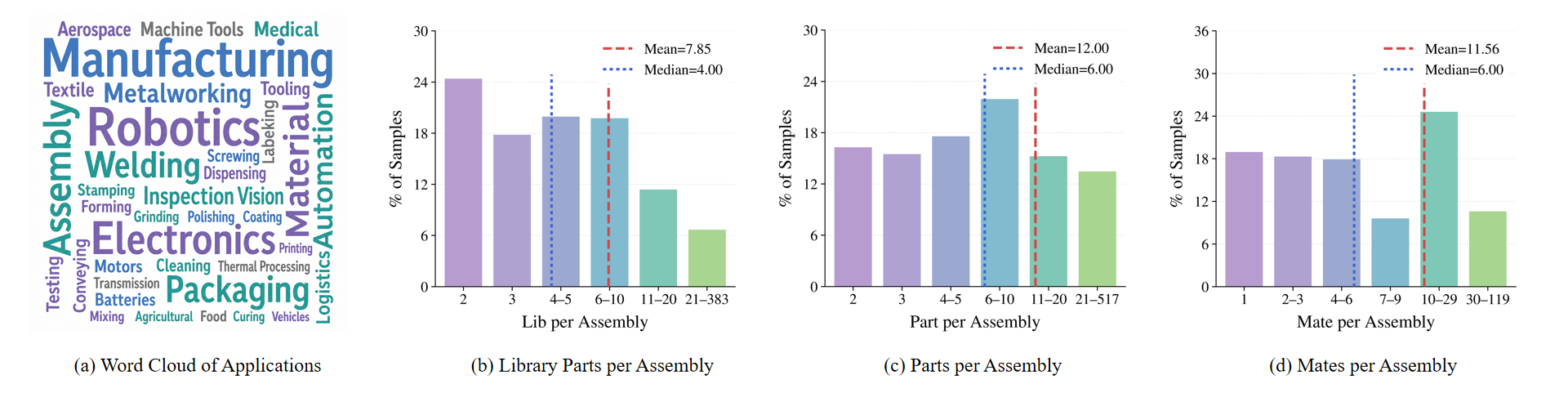}
    \caption{
        Statistical overview of the \name benchmark.
        From left to right, we show
        (a) a word cloud summarizing the application domains,
        (b) the distribution of component instances per assembly,
        (c) the distribution of unique component-library entries per assembly,
        and
        (d) the distribution of mating constraints per assembly.
    }
    \label{fig:assembly_statistics}
\end{figure*}

To address these limitations, we introduce \textbf{\name}, the first large-scale benchmark designed to evaluate vision-language models on 3D mechanical assembly reasoning. As shown in Figure~\ref{fig:assembly_statistics}, \textbf{\name} contains \tai{25\textit{K}} real-world mechanical assemblies, with an average of \tai{12} parts per assembly, spanning diverse industrial domains such as robotics, automotive systems, aerospace equipment, and medical devices.
Each benchmark instance includes a native parametric SolidWorks assembly file (\texttt{.sldasm}) and its associated part files (\texttt{.sldprt}), preserving both CAD editability and the hierarchical relationships among components. To support comprehensive evaluation across geometric and visual representations, we additionally provide B-Rep models (\texttt{.step}), mesh models (\texttt{.obj}), 8-view spherical renderings of the reference 3D assembly, and an assembly feature graph.
Our benchmark evaluates three core capabilities:
\textbf{(i) component identification and 3D spatial reasoning} for selecting the required component instances, including repeated parts, from a candidate library and estimating their 6-DoF positions and orientations from reference assembly images;
\textbf{(ii) part-to-part mating relation reasoning} for inferring the mating relationships, connectivity, and geometric constraints between components;
and \textbf{(iii) tool-augmented agentic assembly reasoning} for adaptively selecting informative viewpoints and iteratively using visualization, measurement, and verification tools to correct component poses, orientations, and mating relationships.

Our main contributions are as follows:
\textbf{(i)} We introduce \textbf{\name}, a large-scale benchmark of \tai{25\textit{K}}  real-world industrial mechanical assemblies, each paired with a candidate 3D part library, a human-engineered ground-truth assembly, multi-view renderings, component-level 6-DoF poses, and explicit part-to-part mate graphs.
\textbf{(ii)} We systematically evaluate state-of-the-art general-purpose LMMs under multiple settings, including component-level 3D spatial reasoning, part-to-part mating reasoning, and tool-augmented agentic assembly reasoning with adaptive viewpoint selection and iterative self-correction.
\textbf{(iii)} We provide a detailed analysis of current models, revealing substantial limitations in selecting repeated component instances, estimating precise positions and orientations, inferring valid mating relationships, avoiding part interpenetration, and scaling to assemblies with large part libraries.

\begin{table}[!t]
\centering
\caption{
Comparison of publicly available 3D assembly datasets.
\#Asm. denotes the total number of assemblies.
Avg. Comp. denotes the average number of component instances per assembly.
\#View denotes the number of reference views.
A green check mark indicates available, while a red cross indicates unavailable.
N/R means not reported.
For GRAPE, \(\dagger\) denotes statistics aggregated from its three
independently reported component catalogs.
For AssemblyBench, \(\ddagger\) denotes the selected reference view,
while the number of candidate isometric views is not reported.
}
\label{tab:assembly_dataset}

\begingroup
\fontsize{6.6}{7.2}\selectfont
\setlength{\tabcolsep}{2.6pt}
\renewcommand{\arraystretch}{1.08}

\resizebox{\columnwidth}{!}{%
\begin{tabular}{
    >{\raggedright\arraybackslash}m{3.15cm}
    >{\centering\arraybackslash}m{0.78cm}
    >{\centering\arraybackslash}m{0.82cm}
    >{\centering\arraybackslash}m{0.62cm}
    >{\centering\arraybackslash}m{0.76cm}
    >{\centering\arraybackslash}m{0.80cm}
    >{\centering\arraybackslash}m{0.60cm}
    >{\centering\arraybackslash}m{0.60cm}
}
\toprule

\multirow{2}{*}{\textbf{Dataset}}
&
\multirow{2}{*}{\textbf{\#Asm.}}
&
\multirow{2}{*}{\makecell{\textbf{Avg.}\\\textbf{Comp.}}}
&
\multirow{2}{*}{\textbf{\#View}}
&
\multirow{2}{*}{\makecell{\textbf{\#Mate}\\\textbf{Type}}}
&
\multirow{2}{*}{\makecell{\textbf{Mate}\\\textbf{Graph}}}
&
\multicolumn{2}{c}{\textbf{3D Object}}
\\

\cmidrule(lr){7-8}

& & & & & &
\textbf{B-Rep}
&
\textbf{Mesh}
\\

\midrule

\makecell[l]{
    \textbf{3D CAD Assembly}\\[-1pt]
    {\fontsize{5.6}{6.0}\selectfont
    \cite{lupinetti2019benchmark}}
}
&
137
&
93.31
&
\xmark
&
5
&
\cmark
&
\cmark
&
\xmark
\\

\specialrule{0.01em}{0.12ex}{0.40ex}

\makecell[l]{
    \textbf{AutoMate}\\[-1pt]
    {\fontsize{5.6}{6.0}\selectfont
    \cite{jones2021automate}}
}
&
92.53K
&
N/R
&
\xmark
&
8
&
\cmark
&
\cmark
&
\xmark
\\

\specialrule{0.01em}{0.12ex}{0.40ex}

\makecell[l]{
    \textbf{JoinABLe}\\[-1pt]
    {\fontsize{5.6}{6.0}\selectfont
    \cite{willis2022joinable}}
}
&
8.25K
&
18.72
&
\xmark
&
7
&
\cmark
&
\cmark
&
\cmark
\\

\specialrule{0.01em}{0.12ex}{0.40ex}

\makecell[l]{
    \textbf{Assemble Them All}\\[-1pt]
    {\fontsize{5.6}{6.0}\selectfont
    \cite{tian2022assemble}}
}
&
12.97K
&
N/R
&
\xmark
&
\xmark
&
\xmark
&
\xmark
&
\cmark
\\

\specialrule{0.01em}{0.12ex}{0.40ex}

\makecell[l]{
    \textbf{GRAPE}\\[-1pt]
    {\fontsize{5.6}{6.0}\selectfont
    \cite{gajek2023grape}}
}
&
35.66K\textsuperscript{\(\dagger\)}
&
10.34\textsuperscript{\(\dagger\)}
&
\xmark
&
\xmark
&
\xmark
&
\xmark
&
\xmark
\\

\specialrule{0.01em}{0.12ex}{0.40ex}

\makecell[l]{
    \textbf{ASAP}\\[-1pt]
    {\fontsize{5.6}{6.0}\selectfont
    \cite{tian2024asap}}
}
&
2.15K
&
N/R
&
\xmark
&
\xmark
&
\xmark
&
\xmark
&
\cmark
\\

\specialrule{0.01em}{0.12ex}{0.40ex}

\makecell[l]{
    \textbf{D4PAS}\\[-1pt]
    {\fontsize{5.6}{6.0}\selectfont
    \cite{zhu2024d4pas}}
}
&
8.67K
&
N/R
&
\xmark
&
\xmark
&
\xmark
&
\xmark
&
\cmark
\\

\specialrule{0.01em}{0.12ex}{0.40ex}

\makecell[l]{
    \textbf{Manual-PA}\\[-1pt]
    {\fontsize{5.6}{6.0}\selectfont
    \cite{zhang2025manualpa}}
}
&
N/R
&
N/R
&
8
&
\xmark
&
\xmark
&
\xmark
&
\cmark
\\

\specialrule{0.01em}{0.12ex}{0.40ex}

\makecell[l]{
    \textbf{AssemblyBench}\\[-1pt]
    {\fontsize{5.6}{6.0}\selectfont
    \cite{li2026assemblybench}}
}
&
2.79K
&
6.7
&
1\textsuperscript{\(\ddagger\)}
&
\xmark
&
\xmark
&
\xmark
&
\cmark
\\

\specialrule{0.08em}{0.25ex}{0.25ex}

\rowcolor{rowblue}
\textbf{OmniCAD (Ours)}
&
\textbf{25K}
&
\textbf{12}
&
\textbf{8}
&
\textbf{21}
&
\textbf{\cmark}
&
\textbf{\cmark}
&
\textbf{\cmark}
\\

\bottomrule
\end{tabular}%
}

\endgroup
\end{table}
\section{Related Work}\label{related_work}

\subsection{Learning to Assemble}
Learning-based 3D part assembly typically predicts the 6-DoF pose of each component to recover a target object. Prior work has progressed from component retrieval and placement~\cite{sung2017complementme,yin2020coalesce} to relational and graph-based pose prediction~\cite{huang2020dynamicgraph,li2020imageassembly,narayan2022rglnet}, with later methods addressing repeated parts, generative assembly, hierarchical structure, assembly order, and visual instruction guidance~\cite{zhang2022instance,cheng2023scorepa,du2024partwhole,xu2024spaformer,zhang2025manualpa,zhao2025assembler}. Mechanical CAD assembly further requires explicit reasoning about connectivity and mating constraints, as explored by AutoMate, JoinABLe, Mates2Motion, and related joint-aware methods~\cite{jones2021automate,willis2022joinable,noeckel2022mates2motion,li2024multijoint}. However, existing approaches largely rely on task-specific models and focus on predefined components, pairwise joints, or category-level objects, whereas our benchmark evaluates general-purpose LMMs on assembly reconstruction from multi-view references and a deduplicated part library, including repeated-part selection, component correspondence, 6-DoF placement, mate-relation inference, and tool-assisted refinement.

\begin{figure*}[h]
    \centering
    \includegraphics[width=\textwidth]{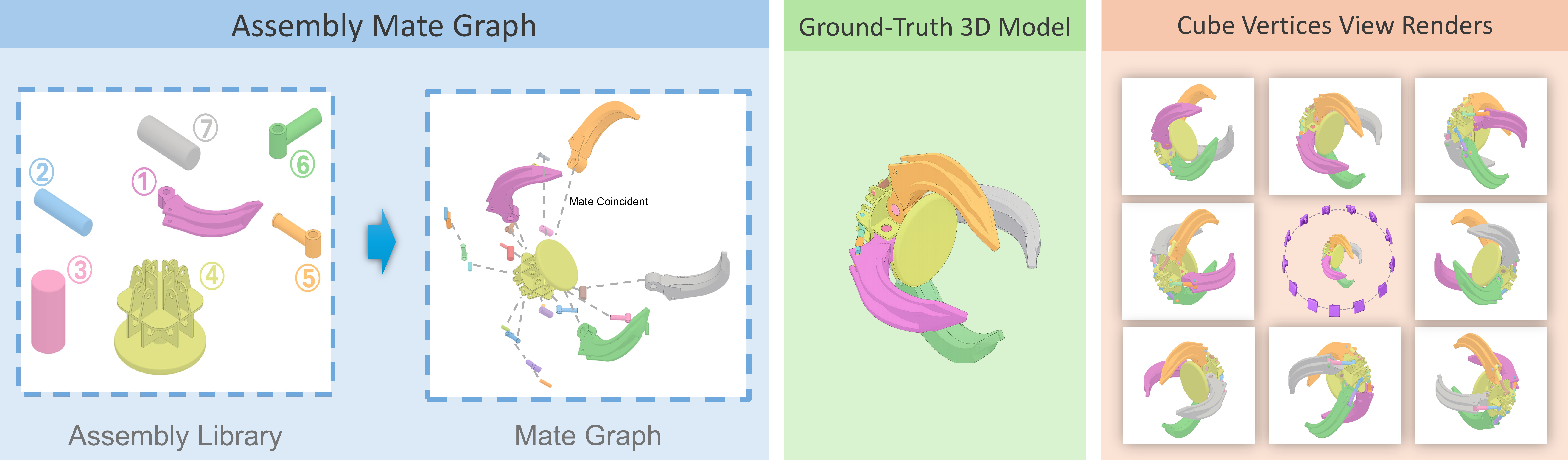}
    \caption{Overview of the \name benchmark with a representative example. From left to right, the figure shows the assembly library and its corresponding mate graph, the ground-truth 3D assembled model, and multi-view renders generated from the vertices of the surrounding view cube.}
    \label{fig:dataset_overview}
\end{figure*}

\subsection{Benchmarks for 3D Part Assembly}
Related 3D datasets have progressed from object-level repositories, such as Princeton Shape Benchmark, ModelNet, and ShapeNet~\cite{shilane2004princeton,wu2015shapenets,chang2015shapenet}, to structured CAD resources, including ABC, DeepCAD, CAD-Coder, and CME-CAD~\cite{koch2019abc,wu2021deepcad,doris2025cadcoder,niu2025cmecad}. However, these CAD datasets model isolated parts and do not capture component instances, assembly poses, or inter-part relations. Part-aware and assembly datasets, such as PartNet, PartNet-Assembly, AutoMate, AssemblyJoint, IKEA-Manual, Breaking Bad, and Assembler~\cite{mo2019partnet,xu2024spaformer,jones2021automate,willis2022joinable,wang2022ikeamanual,sellan2022breakingbad,zhao2025assembler}, introduce part hierarchies, sequential assembly, mating constraints, or visual guidance, but each addresses only a subset of the full problem. No existing benchmark jointly evaluates component-library grounding, repeated-part reuse, 6-DoF placement, component-level mate graphs, geometric conflicts, and tool-assisted iterative refinement for mechanical assemblies.

\section{Benchmark Construction}\label{benchmark}

\textbf{Task Formulation.}
To provide a comprehensive evaluation framework for 3D CAD assembly understanding and prediction, our benchmark formulates the task as recovering a physically plausible assembly from a component library and multimodal assembly evidence. 
Detailed depiction of dataset composition is
arranged in Appendix A due to space constraint.
Given reference assembly views and a library of candidate parts, a model must identify the required component instances, estimate their spatial configurations, infer their assembly relationships, and iteratively refine its predictions using diagnostic tools. Specifically, the benchmark evaluates four capabilities:
\textbf{(i) Component Identification}, requiring models to select the correct component instances from the provided library, including repeated use of the same part when necessary. Because predicted identifiers may differ from ground-truth identifiers, performance is evaluated through semantic and geometric library matching rather than exact string matching.
\textbf{(ii) 3D Pose Reasoning}, requiring models to estimate the position and orientation of each component in the assembly coordinate frame and recover accurate spatial placement, alignment, and relative transformations from visual and geometric evidence.
\textbf{(iii) Mate-Graph Reasoning}, requiring models to reconstruct component-level connectivity and mate types, including concentric, coincident, parallel, perpendicular, distance, and angle constraints.
\textbf{(iv) Agentic Constraint-Aware Assembly Reasoning}, formulating assembly prediction as an iterative process in which models request feedback, such as updated assembly visualizations and geometric conflict checks, and use the resulting diagnostics to refine component poses, mate relations, and physical plausibility.

\textbf{Annotation Curation.}
We construct the {\name} benchmark through a scalable automated annotation pipeline, with full curation details provided in Appendix A. We collect real-world editable 3D CAD assembly models from open-access online repositories, including GrabCAD, 3D ContentCentral, TraceParts, and GitHub~\cite{grabcad,contentcentral,traceparts,github}. The collected projects are implemented in SolidWorks~\cite{solidworks}, a widely adopted industrial CAD platform. Each project consists of a native assembly file (\texttt{.SLDASM}) together with its referenced part files (\texttt{.SLDPRT}), preserving component hierarchy, mating constraints, and geometric dependencies of complete mechanical systems.
Leveraging the SolidWorks API~\cite{solidworks_api}, we systematically extract assembly-level annotations, including component instances, component-library mappings, mate relationships, directed assembly dependency graphs, spatial transformations, and the position and orientation of each component in the assembly coordinate system. To support multimodal assembly reasoning, we further export each design into multiple representations, including component meshes, assembly meshes, rendered reference views, mate graphs, and boundary representations when available. These annotations provide both the input evidence and the ground truth required for evaluating component identification, pose prediction, mate-graph recovery, and physical assembly validity.

\textbf{Statistics of {\name} Benchmark.}
As illustrated in Figure~\ref{fig:assembly_statistics} and Figure~\ref{fig:dataset_overview}, {\name} comprises \tai{25\textit{K}} real-world CAD assembly designs spanning diverse mechanical domains, including robotics, fixtures, mechanisms, electronic housings, and industrial parts. The benchmark contains \tai{300\textit{K}} component instances, \tai{196\textit{K}} unique component-library parts, and \tai{289\textit{K}} mate relations. Each assembly is paired with structured annotations that support both oneshot assembly prediction and iterative agentic assembly reasoning.
Our dataset includes three levels of structured annotations.
\textbf{(i)} \textbf{Component-Level Annotations}, including component IDs, semantic library IDs, mesh paths, component categories, and 3D poses in the assembly coordinate frame.
\textbf{(ii)} \textbf{Assembly-Relation Annotations}, including component-level mate edges, mate types, and directed assembly dependency graphs extracted from native CAD constraints.
\textbf{(iii)} \textbf{Multimodal Reference Evidence}, including component-library geometry, reference assembly views, component-colored rendered views, optional reference assembly OBJ files, and optional mate graphs. These representations enable controlled comparison across different input settings and support the analysis of how geometric, visual, and relational evidence affects assembly prediction.

\section{Evaluation Metrics}

We evaluate one-shot and agentic CAD assembly prediction with metrics covering
component identification, pose accuracy, shape similarity, mate-graph recovery,
physical validity, efficiency, and generation success. For component
identification, \textit{PartID} measures multiset matching accuracy between
predicted components and the ground-truth part-library inventory. \textit{ID-NR}
and \textit{ID-R} report exact inventory reconstruction rates on assemblies
without and with repeated library reuse, respectively.

For pose estimation, predicted instances are aligned to ground-truth instances
within the same library-ID group. \textit{PosErr} is the mean Euclidean
translation error in millimeters. \textit{Pos@10mm} and
\textit{Rot@$10^\circ$} report the fractions of ground-truth components whose
translation and geodesic rotation errors are below $10$ mm and $10^\circ$,
respectively. Missing components are counted as failures in thresholded pose
accuracy.

For shape evaluation, \textit{CD} denotes the symmetric per-component Chamfer
distance computed from transformed component mesh samples. \textit{PA} reports
the fraction of components whose per-component Chamfer distance is below
$10$ mm. \textit{SCD} measures assembly-level shape discrepancy by applying the
same Chamfer distance to the unions of all transformed component point clouds,
i.e., $S_{\mathrm{pred}}=\bigcup_i P_i^{\mathrm{pred}}$ and
$S_{\mathrm{gt}}=\bigcup_i P_i^{\mathrm{gt}}$.

For mate-graph recovery, \textit{PairF1} measures F1 over the mate-edge
multiset, \textit{TypeAcc} evaluates mate-type correctness, and \textit{GSim}
measures directed graph similarity after node alignment. \textit{EntityAcc} is
reported only when comparable mate-entity signatures are available. For
physical validity, \textit{C-Free} reports the fraction of predictions with no
detected geometric conflicts under the diagnostic checker. \textit{RT} is the
mean runtime per evaluated sample; for agentic experiments it includes all
executed interaction rounds and tool-feedback steps. \textit{Avg. Iter.}
reports the mean number of agent rounds. \textit{PR} is the fraction of API attempts that return a parseable, schema-valid prediction JSON.

Metrics directly provided by the input are omitted for the corresponding
setting. Thus, mate-graph-conditioned settings with fixed component inventories
do not report \textit mate-graph-conditioned settings with fixed component inventories
do not report \textit{ID-NR}/\textit{ID-R}, and pose-only mate-graph settings do
not report \textit{PairF1}, \textit{TypeAcc}, or \textit{GSim}.
\section{Experiment and Findings}
\label{sec:experiments}

\begin{figure}[t]
    \centering
    \includegraphics[width=\columnwidth,trim=15 10 20 10,clip]
    {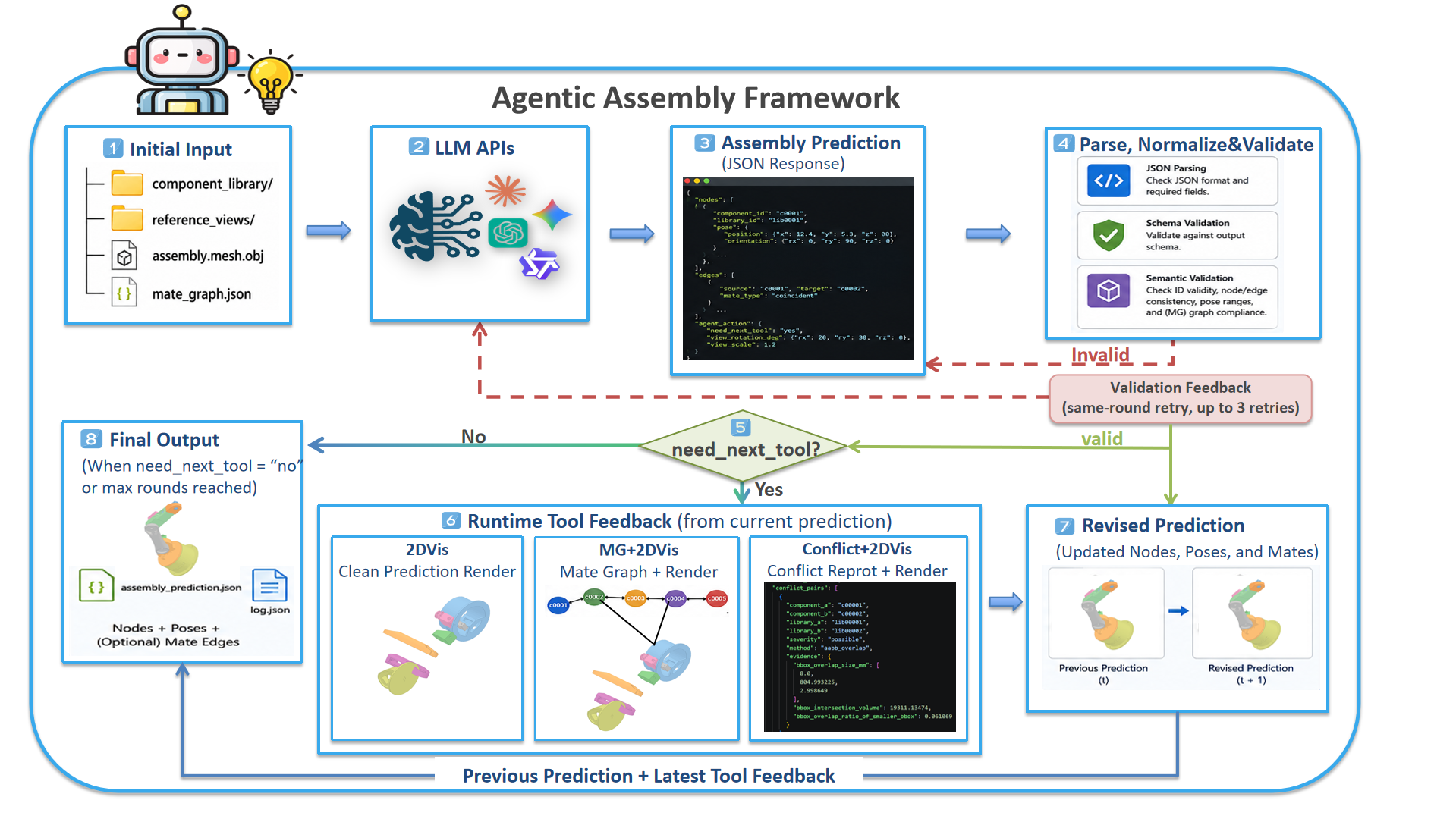}
    \caption{Agentic assembly framework for evaluating LMM tool use in
    mechanical assembly reconstruction.}
    \label{fig:react_framework}
\end{figure}


\begin{table*}[!t]
\centering
\caption{\textbf{One-shot evaluation of LMMs on component-level CAD assembly generation.} Entries marked `--' denote metrics that are not applicable by design:
ID-NR/ID-R are omitted when the input mate graph already specifies the component-instance inventory;
PairF1/TypeAcc/GSim are omitted for pose-only mate-graph-conditioned settings that do not require predicted mate edges. PR is computed per setting as successful final outputs divided by total attempts.}
\label{tab:cad-assembly-oneshot}
\label{tab:cad_oneshot_eval}

\begingroup
\scriptsize
\setlength{\tabcolsep}{2.2pt}
\renewcommand{\arraystretch}{0.62}

\resizebox{\textwidth}{!}{%
\begin{tabular}{l c l cc cccc ccc ccc cc}

\toprule

\multirow{2}{*}{Model}
& \multirow{2}{*}{Size}
& \multirow{2}{*}{Experiment}
& \multicolumn{2}{c}{Component}
& \multicolumn{4}{c}{Pose}
& \multicolumn{3}{c}{Mate Graph}
& \multicolumn{3}{c}{Physical / Shape}
& \multicolumn{2}{c}{Efficiency / Success} \\

\cmidrule(lr){4-5}
\cmidrule(lr){6-9}
\cmidrule(lr){10-12}
\cmidrule(lr){13-15}
\cmidrule(lr){16-17}

& &
& \makecell[c]{ID-NR\\(\%) $\uparrow$}
& \makecell[c]{ID-R\\(\%) $\uparrow$}
& \makecell[c]{Pos@$\tau$\\(\%) $\uparrow$}
& \makecell[c]{Rot@$\tau$\\(\%) $\uparrow$}
& \makecell[c]{CD$_i$\\$\downarrow$}
& \makecell[c]{PA\\(\%) $\uparrow$}
& \makecell[c]{PairF1\\(\%) $\uparrow$}
& \makecell[c]{TypeAcc\\(\%) $\uparrow$}
& \makecell[c]{GSim\\(\%) $\uparrow$}
& \makecell[c]{C-Free\\(\%) $\uparrow$}
& \makecell[c]{SCD\\$\downarrow$}
& \makecell[c]{PosErr\\$\downarrow$}
& \makecell[c]{RT\\$\downarrow$}
& \makecell[c]{PR\\(\%) $\uparrow$} \\
\midrule

\rowcolor{gray!20}
\multicolumn{17}{c}{Commercial Chatbot Systems} \\
\midrule
\multirow{5}{*}{GPT-5.5}  &  \multirow{5}{*}{-}  &  SameColor  &  100.00  &  60.00  &  15.30  &  48.40  &  245.6  &  32.00  &  36.40  &  44.10  &  12.10  &  3.10  &  144.0  &  368.9  &  111.0  &  37.00 \\
    &    &  RefOBJ  &  100.00  &  71.40  &  0.00  &  84.70  &  240.8  &  14.70  &  46.10  &  59.70  &  23.30  &  11.80  &  152.4  &  316.3  &  99.5  &  21.00 \\
    &    &  ColorView  &  92.90  &  57.10  &  15.70  &  43.30  &  349.3  &  25.00  &  35.70  &  44.00  &  8.40  &  10.00  &  232.2  &  481.9  &  171.3  &  35.00 \\
    &    &  MG+Color  &  --  &  --  &  27.90  &  63.20  &  279.7  &  46.70  &  --  &  --  &  --  &  0.00  &  186.5  &  328.6  &  83.1  &  20.00 \\
    &    &  MG+OBJ+Color  &  --  &  --  &  0.00  &  94.10  &  246.9  &  11.10  &  --  &  --  &  --  &  12.50  &  159.6  &  319.6  &  87.4  &  18.00 \\
\midrule
\multirow{5}{*}{GPT-5-mini}  &  \multirow{5}{*}{-}  &  SameColor  &  93.80  &  0.00  &  12.80  &  37.70  &  268.5  &  18.20  &  21.30  &  5.30  &  6.80  &  12.10  &  176.8  &  306.6  &  12.8  &  35.00 \\
    &    &  RefOBJ  &  90.00  &  0.00  &  11.20  &  41.20  &  337.7  &  23.80  &  21.10  &  3.40  &  8.20  &  26.30  &  186.2  &  378.3  &  8.6  &  21.00 \\
    &    &  ColorView  &  93.80  &  15.00  &  11.40  &  36.70  &  305.7  &  16.30  &  22.10  &  6.30  &  5.50  &  11.80  &  204.1  &  351.1  &  12.1  &  36.00 \\
    &    &  MG+Color  &  --  &  --  &  15.30  &  37.50  &  325.8  &  18.50  &  --  &  --  &  --  &  12.10  &  186.0  &  359.9  &  12.0  &  35.00 \\
    &    &  MG+OBJ+Color  &  --  &  --  &  3.90  &  42.90  &  392.1  &  12.90  &  --  &  --  &  --  &  31.20  &  221.7  &  458.8  &  15.0  &  18.00 \\
\midrule
\multirow{5}{*}{Claude Opus 4.8}  &  \multirow{5}{*}{-}  &  SameColor  &  95.00  &  16.10  &  18.60  &  39.30  &  399.4  &  21.90  &  34.80  &  30.00  &  6.30  &  15.80  &  262.7  &  431.4  &  11.8  &  70.31 \\
    &    &  RefOBJ  &  95.30  &  16.10  &  6.20  &  36.10  &  393.1  &  10.70  &  37.80  &  30.00  &  8.10  &  22.80  &  263.6  &  458.7  &  12.6  &  50.60 \\
    &    &  ColorView  &  95.10  &  24.10  &  19.50  &  39.90  &  389.5  &  22.70  &  35.80  &  31.80  &  5.60  &  14.40  &  254.4  &  426.4  &  9.2  &  72.34 \\
    &    &  MG+Color  &  --  &  --  &  23.30  &  44.00  &  375.2  &  26.50  &  --  &  --  &  --  &  11.60  &  240.8  &  407.3  &  14.2  &  70.11 \\
    &    &  MG+OBJ+Color  &  --  &  --  &  9.60  &  39.50  &  372.2  &  13.70  &  --  &  --  &  --  &  16.60  &  239.7  &  418.5  &  16.1  &  53.40 \\
\midrule
\multirow{5}{*}{Gemini 3.1 Pro}  &  \multirow{5}{*}{-}  &  SameColor  &  100.00  &  11.10  &  5.30  &  34.70  &  171.6  &  12.50  &  25.00  &  21.50  &  3.60  &  0.00  &  74.6  &  203.9  &  84.9  &  4.40 \\
    &    &  RefOBJ  &  75.00  &  0.00  &  0.00  &  28.00  &  272.5  &  11.10  &  2.60  &  2.20  &  11.10  &  37.50  &  158.2  &  287.6  &  168.7  &  5.62 \\
    &    &  ColorView  &  96.10  &  6.50  &  8.50  &  36.50  &  365.0  &  20.00  &  17.90  &  13.00  &  12.60  &  6.00  &  228.8  &  426.5  &  49.7  &  32.29 \\
    &    &  MG+Color  &  --  &  --  &  16.70  &  67.00  &  162.7  &  55.00  &  --  &  --  &  --  &  0.00  &  128.8  &  224.3  &  59.0  &  5.00 \\
    &    &  MG+OBJ+Color  &  --  &  --  &  0.00  &  50.00  &  88.2  &  0.00  &  --  &  --  &  --  &  0.00  &  84.2  &  111.9  &  42.6  &  1.50 \\
\midrule
\multirow{5}{*}{Gemini 2.5 Flash-lite}  &  \multirow{5}{*}{-}  &  SameColor  &  95.70  &  8.30  &  8.40  &  35.00  &  253.8  &  16.40  &  22.60  &  13.00  &  8.30  &  2.60  &  160.0  &  277.6  &  1.6  &  39.00 \\
    &    &  RefOBJ  &  96.00  &  20.00  &  9.50  &  34.70  &  272.2  &  16.70  &  13.00  &  5.00  &  14.50  &  2.10  &  177.9  &  298.1  &  30.7  &  18.04 \\
    &    &  ColorView  &  95.70  &  25.20  &  8.10  &  36.70  &  220.2  &  15.50  &  23.70  &  12.50  &  11.00  &  4.30  &  129.0  &  250.0  &  25.4  &  35.05 \\
    &    &  MG+Color  &  --  &  --  &  8.40  &  38.20  &  271.2  &  17.00  &  --  &  --  &  --  &  1.90  &  163.4  &  299.0  &  21.2  &  39.18 \\
    &    &  MG+OBJ+Color  &  --  &  --  &  9.20  &  41.20  &  260.0  &  18.60  &  --  &  --  &  --  &  5.60  &  163.5  &  291.8  &  22.8  &  20.27 \\
\midrule
\midrule
\rowcolor{gray!20}
\multicolumn{17}{c}{Open-Source MLLMs} \\
\midrule
\multirow{5}{*}{LLaMA4 Maverick}  &  \multirow{5}{*}{400B}  &  SameColor  &  95.40  &  16.90  &  21.80  &  47.10  &  602.0  &  27.80  &  41.30  &  20.90  &  13.90  &  18.10  &  405.9  &  647.9  &  18.2  &  22.54 \\
    &    &  RefOBJ  &  97.50  &  5.20  &  19.40  &  45.70  &  576.5  &  26.50  &  38.90  &  19.50  &  13.70  &  23.00  &  421.9  &  545.1  &  203.3  &  36.31 \\
    &    &  ColorView  &  94.60  &  24.40  &  22.20  &  45.90  &  688.7  &  27.60  &  40.70  &  21.00  &  13.50  &  16.20  &  519.9  &  653.8  &  223.9  &  19.91 \\
    &    &  MG+Color  &  --  &  --  &  22.60  &  50.50  &  664.3  &  27.80  &  --  &  --  &  --  &  21.30  &  472.6  &  660.6  &  234.2  &  19.83 \\
    &    &  MG+OBJ+Color  &  --  &  --  &  24.50  &  52.50  &  659.5  &  31.20  &  --  &  --  &  --  &  25.80  &  447.8  &  596.7  &  190.9  &  10.43 \\
\midrule
\multirow{5}{*}{Ministral}  &  \multirow{5}{*}{14B}  &  SameColor  &  98.00  &  48.00  &  17.60  &  45.00  &  697.2  &  25.00  &  45.30  &  26.70  &  16.90  &  29.10  &  488.0  &  655.3  &  88.7  &  13.83 \\
    &    &  RefOBJ  &  97.40  &  90.00  &  12.30  &  52.50  &  164.2  &  20.70  &  37.30  &  14.90  &  15.30  &  4.10  &  140.7  &  195.7  &  571.7  &  1.40 \\
    &    &  ColorView  &  93.00  &  53.20  &  16.10  &  42.80  &  758.7  &  23.40  &  43.10  &  28.50  &  17.30  &  33.50  &  588.2  &  753.3  &  559.1  &  5.71 \\
    &    &  MG+Color  &  --  &  --  &  16.10  &  42.50  &  1860.2  &  20.10  &  --  &  --  &  --  &  32.30  &  1689.3  &  1779.5  &  358.5  &  2.69 \\
    &    &  MG+OBJ+Color  &  --  &  --  &  0.30  &  57.00  &  171.4  &  3.70  &  --  &  --  &  --  &  27.30  &  110.5  &  223.9  &  388.1  &  0.31 \\
\midrule
\multirow{5}{*}{Qwen3.5-9B}  &  \multirow{5}{*}{9B}  &  SameColor  &  100.00  &  33.30  &  7.10  &  25.80  &  180.0  &  7.10  &  35.80  &  18.50  &  20.00  &  28.60  &  101.8  &  210.1  &  263.0  &  1.40 \\
    &    &  RefOBJ  &  100.00  &  0.00  &  2.60  &  39.60  &  390.4  &  2.60  &  30.20  &  20.40  &  9.10  &  9.10  &  223.9  &  493.1  &  302.9  &  2.20 \\
    &    &  ColorView  &  100.00  &  28.60  &  20.20  &  48.00  &  241.7  &  28.10  &  29.00  &  19.40  &  9.30  &  22.20  &  165.9  &  250.7  &  153.8  &  3.60 \\
    &    &  MG+Color  &  --  &  --  &  6.10  &  25.10  &  559.6  &  3.30  &  --  &  --  &  --  &  33.30  &  277.7  &  654.8  &  166.7  &  1.20 \\
    &    &  MG+OBJ+Color  &  --  &  --  &  --  &  --  &  --  &  --  &  --  &  --  &  --  &  --  &  --  &  --  &  --  &  -- \\
\midrule
\multirow{5}{*}{Qwen3.7-Plus}  &  \multirow{5}{*}{-}  &  SameColor  &  94.30  &  1.30  &  14.70  &  34.70  &  380.6  &  17.00  &  20.60  &  13.00  &  8.70  &  4.30  &  238.0  &  416.6  &  288.6  &  51.00 \\
    &    &  RefOBJ  &  98.30  &  3.40  &  7.50  &  37.10  &  290.9  &  9.50  &  26.00  &  19.30  &  11.40  &  12.90  &  188.5  &  334.4  &  112.1  &  23.20 \\
    &    &  ColorView  &  94.40  &  3.30  &  15.00  &  35.10  &  384.7  &  17.10  &  20.80  &  14.50  &  8.30  &  4.70  &  240.6  &  421.2  &  400.8  &  51.60 \\
    &    &  MG+Color  &  --  &  --  &  15.10  &  38.80  &  354.6  &  17.80  &  --  &  --  &  --  &  2.50  &  213.7  &  390.0  &  331.9  &  48.40 \\
    &    &  MG+OBJ+Color  &  --  &  --  &  2.50  &  40.20  &  376.7  &  5.10  &  --  &  --  &  --  &  8.70  &  257.3  &  430.9  &  648.8  &  25.20 \\
\midrule
\bottomrule
\end{tabular}%
}

\endgroup
\end{table*}

\subsection{Experimental Setups}
\label{sec:experimental_setups}

\textbf{Study Setup.}
We evaluate nine proprietary and open-weight LMMs:
GPT-5.5~\cite{openai_gpt55},
GPT-5 mini~\cite{openai_gpt5mini},
Claude Opus 4.8~\cite{anthropic2026claudeopus48},
Gemini 3.1 Pro Preview~\cite{google_gemini3_pro},
Gemini 2.5 Flash-Lite~\cite{google_gemini_25_flash},
Llama 4 Maverick~\cite{meta_llama4},
Ministral 3 14B (v25.12)~\cite{mistralai2025ministral314b},
Qwen3.5-9B~\cite{qwen35_9b}, and
Qwen3.7-Plus~\cite{qwen3_7_plus}.
Each sample uses a deduplicated part library; repeated parts share one
library entry but retain separate instance IDs and poses. Depending on the
setting, models additionally receive multi-view target renderings, a
reference assembly OBJ, or a component-level mate graph. They predict
component instances, component-to-library correspondences, and 6-DoF
poses; without an input mate graph, they also predict component-level mate
relations. We conduct \textbf{(i) Direct One-shot Evaluation}, where the
complete assembly is predicted in one response, and \textbf{(ii) Agentic
Evaluation}, where predictions are iteratively revised using visual and
geometric tool feedback.

\textbf{Direct One-shot Evaluation.}
We evaluate five input configurations to isolate the effects of appearance
cues, explicit assembly geometry, and connectivity priors.
\textit{SameColor} provides the component library and eight target views
rendered from fixed cube-vertex directions with a uniform appearance.
\textit{RefOBJ} provides the component library and the reference assembly
OBJ. \textit{ColorView} provides the component library and eight
color-by-component target views that distinguish component instances.
\textit{MG+Color} augments the color views with the component-level mate
graph, while \textit{MG+OBJ+Color} combines the mate graph, reference
assembly OBJ, and color-by-component views.

\textbf{Implementation of LMMs Agentic Framework Evaluation.}
We evaluate whether iterative tool use can correct errors in component
selection, placement, and assembly structure. At each round, the model
receives the original inputs, its previous structured prediction, and the
latest tool feedback, then returns a revised prediction.
\textit{2DVis} provides a clean rendering of the current prediction under
a specified global view rotation and scale. \textit{MG+2DVis} additionally
provides the component-level mate graph as a structural prior.
\textit{Conflict+2DVis} combines rendered feedback with a geometric
conflict report identifying potentially intersecting component pairs
through AABB-overlap, OBB-overlap, and mesh-interpenetration checks.
All feedback is generated from the current prediction rather than the
ground-truth assembly.

\textbf{Implementation Details.}
All models use the same input serialization, structured prediction schema,
normalization, and evaluation pipeline. Positions are expressed in
millimeters and orientations as XYZ Euler angles in degrees. Predicted
component and library references are validated before evaluation, and
runtime-owned metadata is added after parsing. Agentic runs terminate when
the model no longer requests tool feedback or after ten rounds.


\begin{table*}[!t]
\centering
\caption{\textbf{Agentic evaluation of LMMs on component-level CAD assembly generation.} Entries marked `--' denote metrics that are not applicable by design:
ID-NR/ID-R are omitted when the input mate graph already specifies the component-instance inventory;
PairF1/TypeAcc/GSim are omitted for pose-only mate-graph-conditioned settings that do not require predicted mate edges. PR is computed per setting as successful final outputs divided by total round attempts.}
\label{tab:cad-assembly-agent}
\label{tab:cad_agentic_eval}

\begingroup
\scriptsize
\setlength{\tabcolsep}{1.8pt}
\renewcommand{\arraystretch}{0.62}

\resizebox{\textwidth}{!}{%
\begin{tabular}{l c l cc cccc ccc ccc ccc}

\toprule

\multirow{2}{*}{Model}
& \multirow{2}{*}{Size}
& \multirow{2}{*}{Agent Setting}
& \multicolumn{2}{c}{Component}
& \multicolumn{4}{c}{Pose}
& \multicolumn{3}{c}{Mate Graph}
& \multicolumn{3}{c}{Physical / Shape}
& \multicolumn{3}{c}{Efficiency / Success} \\

\cmidrule(lr){4-5}
\cmidrule(lr){6-9}
\cmidrule(lr){10-12}
\cmidrule(lr){13-15}
\cmidrule(lr){16-18}

& &
& \makecell[c]{ID-NR\\(\%) $\uparrow$}
& \makecell[c]{ID-R\\(\%) $\uparrow$}
& \makecell[c]{Pos@$\tau$\\(\%) $\uparrow$}
& \makecell[c]{Rot@$\tau$\\(\%) $\uparrow$}
& \makecell[c]{CD$_i$\\$\downarrow$}
& \makecell[c]{PA\\(\%) $\uparrow$}
& \makecell[c]{PairF1\\(\%) $\uparrow$}
& \makecell[c]{TypeAcc\\(\%) $\uparrow$}
& \makecell[c]{GSim\\(\%) $\uparrow$}
& \makecell[c]{C-Free\\(\%) $\uparrow$}
& \makecell[c]{SCD\\$\downarrow$}
& \makecell[c]{PosErr\\$\downarrow$}
& \makecell[c]{Step\\$\downarrow$}
& \makecell[c]{RT\\$\downarrow$}
& \makecell[c]{PR\\(\%) $\uparrow$} \\
\midrule

\rowcolor{gray!20}
\multicolumn{18}{c}{Commercial Chatbot Systems} \\
\midrule
\multirow{3}{*}{GPT-5.5}  &  \multirow{3}{*}{-}  &  2DVis  &  98.40  &  54.50  &  18.70  &  46.70  &  182.5  &  30.30  &  32.70  &  43.40  &  15.50  &  16.00  &  120.6  &  224.7  &  1.41  &  148.0  &  62.00 \\
  &    &  MG+2DVis  &  --  &  --  &  21.20  &  50.10  &  189.5  &  33.30  &  --  &  --  &  --  &  20.00  &  135.6  &  246.0  &  1.14  &  130.9  &  53.74 \\
  &    &  Conflict+2DVis  &  100.00  &  46.30  &  17.50  &  42.70  &  227.9  &  28.20  &  30.50  &  33.40  &  14.60  &  20.70  &  169.1  &  291.6  &  2.18  &  218.9  &  70.96 \\
\midrule
\multirow{3}{*}{GPT-5-mini}  &  \multirow{3}{*}{-}  &  2DVis  &  98.10  &  13.90  &  18.20  &  44.90  &  209.5  &  27.20  &  28.60  &  11.90  &  13.00  &  27.30  &  124.6  &  230.3  &  1.28  &  22.0  &  50.22 \\
  &    &  MG+2DVis  &  --  &  --  &  18.30  &  48.30  &  163.4  &  29.70  &  --  &  --  &  --  &  20.90  &  102.4  &  187.0  &  1.27  &  20.5  &  44.34 \\
  &    &  Conflict+2DVis  &  100.00  &  11.10  &  16.20  &  47.30  &  232.4  &  27.20  &  31.80  &  13.60  &  15.10  &  27.70  &  151.3  &  270.5  &  1.64  &  32.7  &  49.21 \\
\midrule
\multirow{3}{*}{Claude Opus 4.8}  &  \multirow{3}{*}{-}  &  2DVis  &  98.60  &  23.20  &  15.70  &  37.70  &  874.6  &  18.70  &  35.00  &  36.70  &  6.10  &  13.70  &  702.2  &  875.0  &  2.69  &  55.1  &  87.07 \\
  &    &  MG+2DVis  &  --  &  --  &  16.70  &  38.90  &  868.0  &  20.80  &  --  &  --  &  --  &  13.40  &  716.7  &  893.4  &  3.17  &  60.0  &  87.82 \\
  &    &  Conflict+2DVis  &  98.60  &  21.70  &  15.00  &  36.80  &  347.6  &  19.60  &  35.10  &  32.80  &  5.40  &  17.30  &  216.2  &  397.0  &  2.13  &  44.0  &  84.25 \\
\midrule
\multirow{3}{*}{Gemini 3.1 Pro}  &  \multirow{3}{*}{-}  &  2DVis  &  96.80  &  33.30  &  21.60  &  54.30  &  163.3  &  37.00  &  42.30  &  33.80  &  21.90  &  14.30  &  92.5  &  195.3  &  1.59  &  241.2  &  25.48 \\
  &    &  MG+2DVis  &  --  &  --  &  30.70  &  59.70  &  174.3  &  48.90  &  --  &  --  &  --  &  15.00  &  100.1  &  227.5  &  1.67  &  255.9  &  32.21 \\
  &    &  Conflict+2DVis  &  100.00  &  100.00  &  34.60  &  66.00  &  85.4  &  57.10  &  42.60  &  44.70  &  16.00  &  23.10  &  39.9  &  103.8  &  1.42  &  283.4  &  11.59 \\
\midrule
\multirow{3}{*}{Gemini 2.5 Flash-lite}  &  \multirow{3}{*}{-}  &  2DVis  &  100.00  &  42.90  &  22.70  &  48.80  &  187.0  &  37.80  &  19.60  &  15.80  &  23.10  &  16.00  &  101.9  &  221.2  &  3.85  &  46.4  &  36.50 \\
  &    &  MG+2DVis  &  --  &  --  &  21.20  &  56.50  &  141.6  &  29.30  &  --  &  --  &  --  &  9.10  &  96.5  &  184.7  &  1.05  &  38.1  &  17.19 \\
  &    &  Conflict+2DVis  &  100.00  &  25.00  &  17.70  &  65.20  &  104.7  &  30.40  &  34.20  &  25.10  &  14.10  &  43.80  &  79.7  &  118.4  &  1.88  &  66.5  &  10.92 \\
\midrule
\midrule
\rowcolor{gray!20}
\multicolumn{18}{c}{Open-Source MLLMs} \\
\midrule
\multirow{3}{*}{LLaMA4 Maverick}  &  \multirow{3}{*}{400B}  &  2DVis  &  100.00  &  25.00  &  26.50  &  55.30  &  280.6  &  28.20  &  36.20  &  20.40  &  15.80  &  11.40  &  183.3  &  179.3  &  1.00  &  42.8  &  17.50 \\
  &    &  MG+2DVis  &  --  &  --  &  21.30  &  53.00  &  182.9  &  33.50  &  --  &  --  &  --  &  18.60  &  143.9  &  200.6  &  1.00  &  51.5  &  22.50 \\
  &    &  Conflict+2DVis  &  96.00  &  16.70  &  32.80  &  61.80  &  278.3  &  36.10  &  40.20  &  24.90  &  22.30  &  19.40  &  166.6  &  160.2  &  2.29  &  67.9  &  30.74 \\
\midrule
\multirow{3}{*}{Ministral}  &  \multirow{3}{*}{14B}  &  2DVis  &  100.00  &  37.50  &  20.20  &  51.20  &  213.0  &  32.50  &  41.30  &  18.30  &  15.30  &  37.00  &  166.6  &  238.8  &  1.36  &  100.1  &  18.10 \\
  &    &  MG+2DVis  &  --  &  --  &  19.60  &  47.40  &  231.7  &  28.50  &  --  &  --  &  --  &  40.70  &  164.1  &  266.7  &  2.71  &  96.4  &  30.65 \\
  &    &  Conflict+2DVis  &  100.00  &  62.50  &  24.50  &  52.50  &  234.8  &  27.30  &  40.60  &  18.30  &  13.20  &  29.60  &  163.0  &  270.9  &  2.54  &  94.6  &  29.22 \\
\midrule
\multirow{3}{*}{Qwen3.5-9B}  &  \multirow{3}{*}{9B}  &  2DVis  &  92.90  &  28.60  &  19.20  &  52.50  &  207.6  &  24.30  &  28.80  &  27.00  &  17.30  &  19.00  &  153.7  &  190.2  &  1.00  &  150.3  &  10.50 \\
  &    &  MG+2DVis  &  --  &  --  &  16.70  &  55.80  &  71.7  &  32.50  &  --  &  --  &  --  &  40.00  &  44.2  &  100.5  &  1.00  &  867.1  &  2.50 \\
  &    &  Conflict+2DVis  &  88.90  &  40.00  &  26.20  &  49.30  &  188.0  &  30.90  &  36.40  &  29.40  &  19.50  &  35.70  &  119.1  &  202.8  &  1.00  &  189.0  &  7.00 \\
\midrule
\multirow{3}{*}{Qwen3.7-Plus}  &  \multirow{3}{*}{-}  &  2DVis  &  97.90  &  17.60  &  21.90  &  54.80  &  613.1  &  29.50  &  33.70  &  36.20  &  21.40  &  24.60  &  378.9  &  633.9  &  2.03  &  272.9  &  49.44 \\
  &    &  MG+2DVis  &  --  &  --  &  29.30  &  66.50  &  260.1  &  37.40  &  --  &  --  &  --  &  27.90  &  234.5  &  276.0  &  3.88  &  436.6  &  51.54 \\
  &    &  Conflict+2DVis  &  97.60  &  14.30  &  24.70  &  55.50  &  899.7  &  31.20  &  36.20  &  33.60  &  24.50  &  23.20  &  535.7  &  990.4  &  2.12  &  300.4  &  45.25 \\
\midrule
\bottomrule
\end{tabular}%
}

\endgroup
\vspace{-15pt}
\end{table*}

\subsection{Main Results and Findings}
\label{sec:main_results}

Results of the direct one-shot and agentic studies are shown in
Table~\ref{tab:cad_oneshot_eval} and
Table~\ref{tab:cad_agentic_eval}, respectively. Current LMMs identify
non-repeated components and estimate coarse orientations relatively well,
but struggle with repeated-part reuse, precise translation, complete mate
relations, and collision-free placement. Agentic feedback corrects some
large spatial errors but provides limited fine-grained alignment gains.

\noindent\textbf{$\bullet$ Component Identification.}
Performance degrades sharply when a library part is reused. Under
ColorView, average ID-NR is 95.1\%, compared with only 26.4\% ID-R,
indicating that repeated-instance assignment remains substantially harder
than one-to-one correspondence.

\noindent\textbf{$\bullet$ Component Placement.}
Translation is substantially more difficult than orientation estimation.
The best Pos@10 is 27.9\% in the one-shot study and 34.6\% in the agentic
study, while rotation accuracy is consistently higher. With
$\tau_{\mathrm{CD}}=10$ mm, the best PA increases only from 55.0\% to
57.1\%, indicating weak metric localization and precise alignment.

\noindent\textbf{$\bullet$ Component-Level Mate Reasoning.}
In the one-shot study, the best PairF1, TypeAcc, and GSim are 46.1\%,
59.7\%, and 23.3\%, respectively; the agentic study reaches 42.6\%
PairF1 and 24.5\% GSim. The low graph similarity indicates missing or
incorrect edges and erroneous mate assignments to repeated instances.

\noindent\textbf{$\bullet$ Impact of Visual and Geometric Inputs.}
Color-by-component views improve repeated-part identification for six of
the nine models relative to same-color views, but gains in pose and shape
accuracy are inconsistent. Adding the reference assembly OBJ to MG+Color
improves Pos@10 for only two of the eight comparable models, indicating
that current LMMs cannot reliably associate the assembly mesh with library
parts and recover their transformations.

\noindent\textbf{$\bullet$ Impact of Mate-Graph Conditioning.}
Compared with ColorView, MG+Color improves both Pos@10 and Rot@10 for
seven of the nine models. The mate graph specifies component connectivity
and reduces relational uncertainty, but does not provide the relative
translations and rotations required for precise placement.

\noindent\textbf{$\bullet$ Agentic Visual and Conflict-Guided Reasoning.}
Compared with direct ColorView generation, 2D visual feedback improves
Pos@10 for seven models, Rot@10 for eight, and GSim for eight. Conflict
feedback improves the conflict-free rate for seven models and raises the
best C-Free to 43.8\%, while the best agentic Pos@10 and PA reach 34.6\%
and 57.1\%. However, more than half of the predictions still contain
conflicts, showing that these tools mainly support coarse correction.

\noindent\textbf{$\bullet$ Generation Reliability and Efficiency.}
Peak API SR reaches 72.3\% in the one-shot study and 71.5\% in the
agentic study, while several settings remain below 10\%. Gemini 3.1 Pro
achieves the best one-shot PA of 55.0\% under MG+Color with only 0.3\%
API SR, so geometric quality must be interpreted jointly with generation
coverage. Its runtime increases from 49.7 s under direct ColorView to
283.4 s under Conflict+2DVis, showing that iterative feedback remains
costly.
\section{Conclusion}
We introduce \textit{\name}, a large-scale benchmark for evaluating LMMs on 3D mechanical assembly reasoning, comprising \tai{25\textit{K}} real-world assemblies with deduplicated part libraries, engineer-authored ground-truth assemblies, multi-view renderings, component-level 6-DoF poses, and explicit mate graphs. We evaluate state-of-the-art LMMs under both one-shot and tool-augmented agentic settings across component identification, pose estimation, mating-relation inference, physical validity, and iterative refinement. Our analysis reveals critical limitations, including unreliable reuse of repeated parts, inaccurate component positions and orientations, incorrect or incomplete mate graphs, severe part interpenetration, and poor scalability to assemblies with large component libraries. By exposing these failure modes at scale, \textit{\name} provides a standardized foundation for developing more reliable, physically grounded, and scalable assembly-reasoning systems.

\bibliography{aaai2027}

\end{document}